%% file: main.tex
\documentclass[10pt,twocolumn,letterpaper]{article}

\usepackage{iccv}      


\definecolor{iccvblue}{rgb}{0.21,0.49,0.74}
\usepackage[pagebackref,breaklinks,colorlinks,allcolors=iccvblue]{hyperref}

\usepackage{arydshln}

\def\paperID{15622} 
\def\confName{ICCV}
\def\confYear{2025}

\title{NGD: Neural Gradient Based Deformation for Monocular Garment Reconstruction}

\author{
Soham Dasgupta \quad
Shanthika Naik \quad
Preet Savalia \quad
Sujay Kumar Ingle \quad
Avinash Sharma \\
Indian Institute of Technology Jodhpur \\
{\tt\small \{sohamd, shanthikanaik, b22ai036, d23csa003, avinashsharma\}@iitj.ac.in}
}

\begin{document}
\twocolumn[{
    \renewcommand\twocolumn[1][]{#1}
    \maketitle
    \input{images/fig_teaser}
    }]

\input{sec/0_abstract}    
\input{sec/1_intro}

\input{sec/2_RelatedWorks}

\input{sec/3_Methodology}

\input{sec/4_Experiments}

\input{sec/5_Ablations}
\input{sec/6_Conclusion}

\bibliographystyle{ieeenat_fullname}
\bibliography{main}


\end{document}

%% file: images/fig_teaser.tex

\begin{center}
    \includegraphics[width=\textwidth]{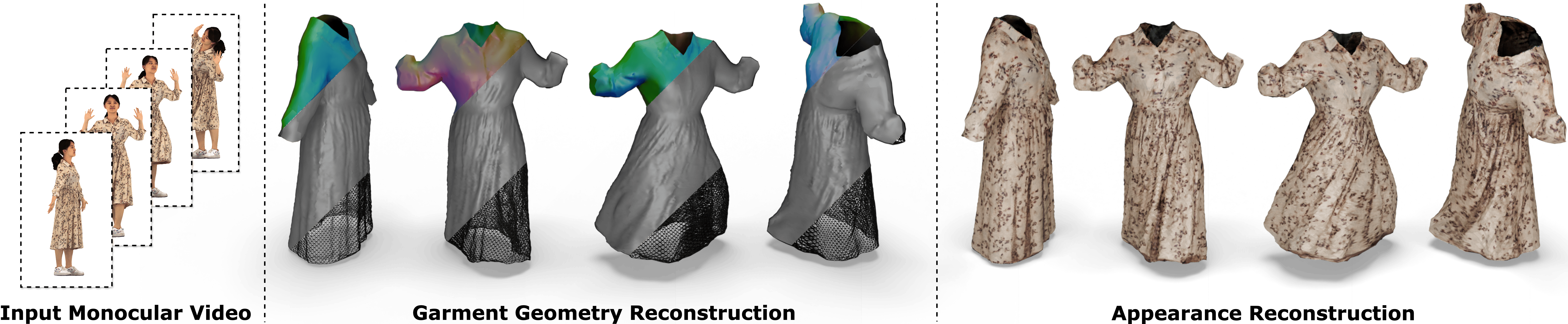}
    \captionof{figure}{Our method reconstructs high-fidelity garment geometry and appearance from input monocular video.}
    \label{fig:teaser}
\end{center}

%% file: sec/0_abstract.tex
\begin{abstract}
Dynamic garment reconstruction from monocular video is an important yet challenging task due to the complex dynamics and unconstrained nature of the garments. Recent advancements in neural rendering have enabled high-quality geometric reconstruction with image/video supervision. However, implicit representation methods that use volume rendering often provide smooth geometry and fail to model high-frequency details. While template reconstruction methods model explicit geometry, they use vertex displacement for deformation which results in artifacts. Addressing these limitations, we propose NGD, a \textbf{N}eural \textbf{G}radient-based \textbf{D}eformation method to reconstruct dynamically evolving textured garments from monocular videos. Additionally, we propose a novel adaptive remeshing strategy for modeling dynamically evolving surfaces like wrinkles and pleats of the skirt, leading to high-quality reconstruction. Finally, we learn dynamic texture maps to capture per-frame lighting and shadow effects. We provide extensive qualitative and quantitative evaluations to demonstrate significant improvements over existing SOTA methods and provide high-quality garment reconstructions.

\end{abstract}

%% file: sec/1_intro.tex
\section{Introduction}
\label{sec:intro}


Recent advances in computer vision have enabled large-scale digitization of 3D garments for immersive AR/VR platforms, revolutionizing Social Media, E-Commerce, Gaming, and Entertainment industries. The sheer diversity, complex dynamics, and intricate articulations make garment digitization and modeling significantly challenging. Unlike conventional digital garment creation methods involving artists, which demanded expertise, time, and labor, deep learning has enabled garment digitization from images and videos \cite{GaussianGarments, physavatars, sewformer, wordrope, recmv}. Multi-view video inputs are used to obtain high-quality garment digitization, but they often require expensive calibrated multi-camera setups \cite{D3GA, GaussianGarments, physavatars, diffavatar, selfrecon}, and hence difficult to scale. In comparison, monocular video inputs are easy to acquire and scalable with an abundance of ``in the wild" videos available. Nevertheless, garment digitization from monocular video needs to reconstruct the dynamically evolving garment geometry and appearance while addressing the classical challenges like modeling varying garment sizes, non-rigid deformations due to body shapes and poses, and the diverse topology of garments.

Advancements in differentiable rendering have made it possible to achieve high-quality geometry reconstruction from monocular videos. \cite{GaussianGarments,scarf,recmv, dgarments,DeepCap,pergamo}. The existing approaches for garment reconstruction can be divided into implicit surface deformation methods \cite{scarf, recmv} and explicit template deformation methods \cite{dgarments, pergamo}. SCARF \cite{scarf} is one of the first works to use implicit surface representation using Neural Radiance Fields (NeRF)\cite{nerf}; nevertheless, the geometric quality is limited by constraints inherent to volume rendering approaches. REC-MV \cite{recmv} addresses this limitation by optimizing for both explicit feature curves and implicit garment surfaces. However, the use of implicit representations adds an overhead of surface extraction and the resulting surface is smooth, losing out high-fidelity surface details. Pergamo \cite{pergamo} and DGarments \cite{dgarments} deform garment templates using vertex displacements, with deformation supervised by differentiable rendering and SMPL-interpolated skinning weights. However, these methods rely on a fixed template constraining its ability to model dynamically varying topology, like the pleats of the skirt. Additionally, direct vertex displacement via differentiable rendering causes abrupt, sharp local changes and requires additional regularisers for smoothening undesirable local deformations. This often results in an over-smoothed surface and fails to capture high-frequency details.


To address the above limitations, we develop a \textit{Neural Gradient Based Deformation} method to reconstruct dynamic garments from input monocular video. Our method models appearance and geometry separately, as learning them together might result in appearance being corrected to compensate for geometric inaccuracies and vice versa. We propose a novel deformation parameterization that decomposes surface deformations into a frame-invariant component representing the global shape and a frame-dependent component modeling the pose-specific local surface deformations of the base garment mesh. Specifically, our deformation parameterization adopts NJF \cite{NJF} to model garment reconstruction, which learns a local Jacobian field defined on the garment surface, followed by a Poisson solve to predict the global garment deformation in canonical space. This addresses the aforementioned limitation of existing template-based methods, which directly model deformation at the vertex level, often resulting in jagged surfaces when applied without any surface smoothness regularizer \cite{icmr}. These canonical garments are skinned to model garment reconstruction to map to the corresponding input monocular views and optimized via a differentiable renderer \cite{nvdiffrast}. While there are existing approaches that leverage NJF \cite{NJF} to model deformation with a differentiable renderer, \cite{textdeformer}, we develop a gradient-based deformation approach to model temporal deformations from the monocular input video. Unlike existing methods that directly optimize with colored images, which can result in inaccuracies due to ambiguities between shadows and textures, we use \textit{diffuse garment images}. Additionally, we design an adaptive remeshing strategy to iteratively increase the mesh resolution in the regions of high-frequency geometrical details. This enables regions with fine details to be modeled by higher mesh resolutions and also freely deform the template to model extremely loose garments. Finally, we learn appearance via dynamic texture maps at each frame to capture lighting and shadow effects. ~\autoref{fig:teaser} visualizes the high-fidelity dynamic textured garment reconstructed by our method from an input monocular video. 
In summary, our key technical contributions are as follows:
\begin{itemize}
    \item We propose a novel method to reconstruct dynamically evolving textured garments from monocular videos. 
    \item Our novel deformation parameterization, combined with the novel adaptive remeshing, enables modeling extremely loose garments with high-frequency details.
    \item We provide qualitative and quantitative comparisons with existing methods to show significant improvements, especially on loose garments.
\end{itemize}

\noindent \textbf{Code: }\href{https://github.com/astonishingwolf/NGD/}{https://github.com/astonishingwolf/NGD/}.

%% file: sec/2_RelatedWorks.tex
\input{images/fig_overview}

\input{images/fig_architecture}

\section{Related Works}
\label{sec:related_works}


A large number of existing methods attempted clothed human reconstruction from single or multi-view images ~\cite{pifuhd, pamir, ECON, ARCH, arch++, sith, sifu} or videos ~\cite{PHORHUM, MonoClothCap, icon, scanimate, neural_body, selfrecon, vid2avatar, reloo}, albeit cannot extract garment mesh separately.
On the other hand, several existing garment reconstruction methods \cite{bcnet, reef, clothwild, implicitPCA, dig, drapenet, sewformer, li2024_garmentrecovery, smplicit, layernet} recover garments from monocular image. However, these single-image reconstruction methods require supervised training on a large dataset.  Please refer to the supplementary for a detailed discussion of these methods.

Multiview images can recover garments in a self-supervised manner. Diffavatar \cite{diffavatar} uses sewing patterns to represent garments and obtain simulation-ready garments from multiview images. Gaussian Garments \cite{GaussianGarments} combines physics simulation with Gaussian splats~\cite{gaussianSplatting} to obtain physically plausible garments from multiview inputs. The rendering captures fine details down to the level of furs. While multiview reconstruction provides rich garment digitization solutions, the multiview camera setups are generally expensive, hence monocular videos provide a cheap, scalable alternative.

DeepCap \cite{DeepCap} is one of the pioneering approaches to reconstructing loose garments from monocular video. However, it considers the first frame as a template, requiring expensive preprocessing including 3D scanning of a clothed human, segmentation, and reconstruction of the garment and human separately. Methods like Pergamo \cite{pergamo} deform garment templates using SMPL-interpolated skinning weights, followed by rendering loss optimization. \cite{scarf} integrates a parametric body model with \cite{nerf} representation for garment reconstruction; however, the geometric quality is constrained by NeRF’s inherent limitations. REC-MV \cite{recmv} uses implicit-explicit representation to achieve geometrically consistent and temporally coherent garment reconstruction. Despite this, they lack detailed textures, and their use of initial implicit garment representations leads to smoothing effects, compromising high-fidelity detail. The recent method, DGarments \cite{dgarments}, achieves state-of-the-art performance in geometry reconstruction from monocular video by introducing a multi-hypothesis deformation module. However, they fail to large deformations and struggle with loose clothing.


%% file: images/fig_overview.tex
\begin{figure}[t]
  \centering
   \includegraphics[width=\linewidth]{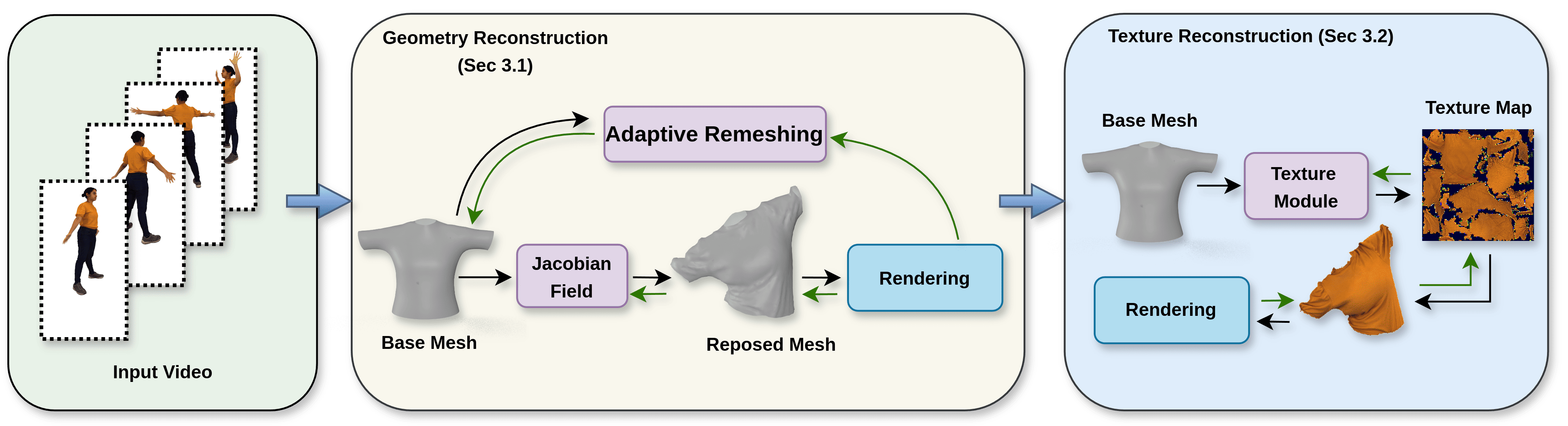}

   \caption{\textbf{Method overview:} Given an input video, we reconstruct dynamically evolving textured garment meshes using our Geometry and Appearance Reconstruction module. }

\label{fig:overview}
\end{figure}

%% file: images/fig_architecture.tex
\begin{figure*}
  \centering
    \includegraphics[width=\linewidth]{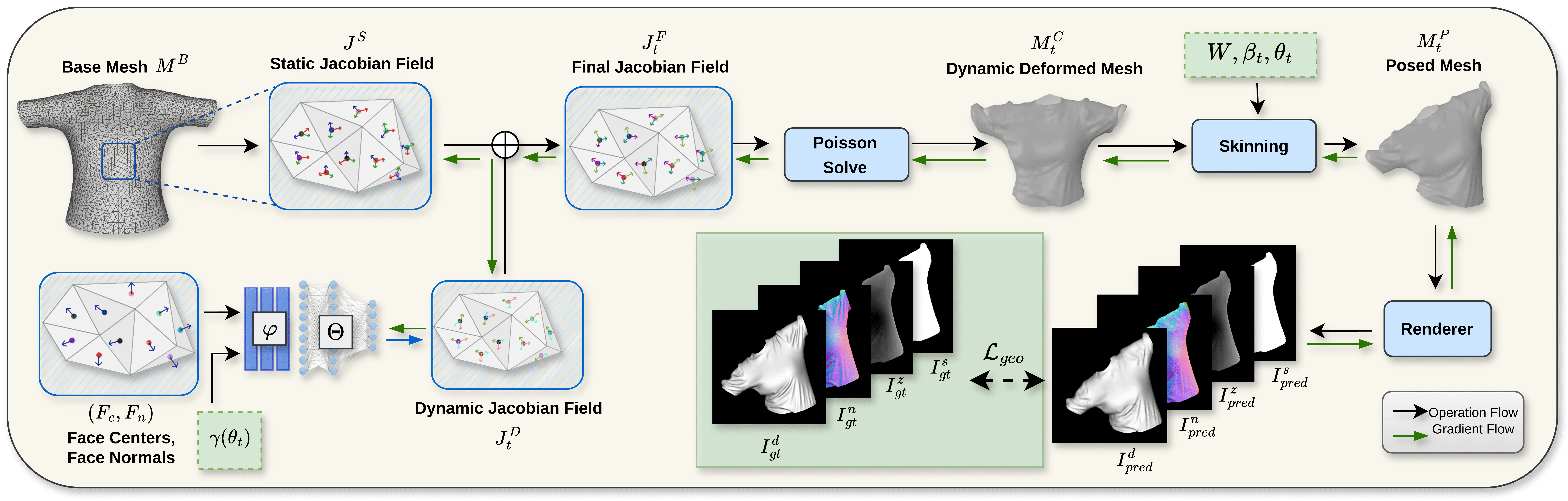}
    \caption{In Geometry Reconstruction Module we introduce a novel deformation parameterization to deform a base mesh $M^B$ to desired target mesh via learning a Jacobian Field guided by differentiable rendering supervision from input monocular video.}
    \label{fig:geometry_archi}
\end{figure*}

%% file: sec/3_Methodology.tex

\section{Method}
\label{sec:formatting}

We present NGD, a novel approach for reconstructing dynamically evolving textured garment meshes from given input monocular video. Our method is composed of geometry and appearance reconstruction modules, as shown in ~\autoref{fig:overview}. As part of our geometry reconstruction module, we introduce a novel deformation parameterization over a base garment mesh to accurately capture and aggregate garment deformations across input frames. This parameterization decomposes deformations into a frame-invariant component representing the global canonical shape and a frame-dependent component modeling the pose-specific local surface deformations of the garment. To further improve geometric fidelity, we also propose a novel Gradient-Based Remeshing Strategy ~\autoref{sec:remeshing}, which adaptively refines the mesh resolution in regions exhibiting high curvature thereby facilitating the precise modeling of intricate details, such as wrinkles and folds. Our appearance reconstruction module ~\autoref{subsec:texture_map} learns garment appearance by learning a frame-invariant base texture map and a frame-dependent dynamic texture map that captures the visual characteristics of the garments.

\input{images/fig_remeshing}

\subsection{Geometric Reconstruction Module}
\label{sec:geometric_module}
The base garment mesh $M^B$ is a 2-manifold embedded in 3D Euclidean space $\mathbb{R}^3$. Let $\mathbf{V}:=\{v_i \in \mathbb{R}^{3}\}_{i = 0}^{N}$, $\mathbf{F}:=\{f_j \in \mathbb{N}^{3}\}_{j = 0}^{M}$ and $\mathbf{E}:=\{e_l \in \mathbb{N}^{2}\}_{l = 0}^{L}$ be the vertices, faces and edges of the mesh $M^B$ respectively. We separately model the global deformations capturing garment-specific design features (such as collars, and necklines) as well the local dynamic deformations (such as wrinkles) on $M^B$ in T-pose at every time-frame. To achieve this, we find a mapping function $\Phi_t$ that transforms the base mesh $M^B$ to a desired mesh $\tilde{M}_t$ in canonical space (T-pose) that captures these dynamic deformations at each time-frame $t$.
This mapping function $\Phi_t:\mathbb{R}^{N \times 3} \to \mathbb{R}^{N \times 3}$ is approximated by optimizing for Jacobian fields and using Poisson solve to obtain deformed mesh vertices \cite{NJF}. However, TextDeformer \cite{textdeformer}, which optimizes for a single static mesh, we need to approximate a temporally varying mapping function $\Phi_t$ corresponding to every frame.


Thus, given input video frames $\mathbf{I}=\{I_t\}_{t = 0}^{T}$, the goal is to find the optimal deformation function $\Phi_t$ at every frame by solving the following equation in the least square form:
\begin{equation}
    \Phi_t^* = \min_{\Phi_t} \sum_{f_j \in F} |f_j| \, \| \nabla_i \Phi_t - J_j \|^2,
\label{eq:possion}
\end{equation}


where $\nabla $ is the gradient operator. The function $\Phi_t^*$ ideally maps the base garment $M^B$ to target mesh $\tilde{M}_t$. The solution to the above equation \autoref{eq:possion} is obtained by solving a Poisson system \cite{NJF}. This mapping function $\Phi_t^*$ is indirectly estimated by optimizing for the Jacobians $J_t^F$ to obtain the canonical mesh $M_t^C$, which is the closest approximation of the desired mesh $\tilde{M}_t$.

\noindent \textbf{Intrinsic Deformation Fields:}
Building on the aforementioned Jacobian Field formulation, we propose a novel deformation parameterization for dynamic garment modeling by splitting $J_t^F$ into two sub-fields, a frame-invariant static Jacobian Field $J^S \in \mathbb{R}^{M \times 3 \times 3}$, and a frame-specific dynamic Jacobian Field $J^D_t \in \mathbb{R}^{M \times 3 \times 3}$. $J^S$ captures the global garment shape specific to the input video garment style. This static field is defined at each face center of the base mesh and is optimized directly across all frames. The dynamic field $J_t^D$ captures the pose-specific surface deformation at each image frame and is predicted by a neural network $f_G$.

The \autoref{fig:geometry_archi} shows how these two Jacobian Fields model per-frame deformations in the canonical space.
The neural network $f_G=f_{\Theta} \circ f_{\varphi}$ is composed of hash-grid encoder $f_{\varphi}$ and an MLP $f_{\Theta}$. At every time-step $t$, we use face centers $F_C$, face normals $F_N$ of the reposed static canonical garment mesh, and pose information for conditioning the neural network. Conditioning on the pose defined by joint angles $\theta_t$ helps prevent overfitting to the input view and enables better generalization of deformations across frames. We use the PCA (Principle Component Analysis) for encoding the pose parameters as $\gamma(\theta_t)$. More details about the pose encoding are provided in the Suppl. Finally, the MLP takes as input latent encoding of $F_N$, $F_C$ and $\gamma(\theta_t)$ from $f_{\varphi}$ to predict $J^D_t$. The final Jacobian field is defined as $J^F_t=J^S+J^D_t$. The final Jacobian field $J_t^F$ is solved via the Poisson system to obtain the canonical garment $M^C_t$ encompassing both global garments specific as well as local surface deformations. 

~\\
\noindent \textbf{Skinning Transformation:}
The canonical garment $M_t^C$ is subsequently skinned to obtain the reposed garment for every time-frame $M^P_t$ defined as follows:
\begin{equation}
    M^P_t = S(M^C_t,\beta_t,\theta_t,W),
\end{equation}
where $S(.)$ is the skinning function, $\beta_t$ and $\theta_t$ are the shape and pose parameters, and $W$ is the garment skinning weights. 
This reposed mesh is rendered to obtain diffuse and depth images of the garment. 
The pseudo ground truth extracted from the input images guides the optimization of the Jacobian Field $J^S$ and the neural network $f_G$ parameters via a differentiable renderer. \autoref{fig:geometry_archi} provides a visual overview of our geometry reconstruction module.

~\\
 \noindent \textbf{Local Minima:} The optimization is often trapped in local minima while minimizing the local rendering losses. To address this, we introduce a novel exponentially decaying noise applied to the vertices of the final skinned mesh iteratively. This noise encourages the model to prioritize global geometry in the initial iterations, preventing early overfitting to local details.  This heuristic adds no computational overhead while significantly improving the reconstruction quality of loose garments (refer to supplementary for detailed discussion).

~\\
\noindent \textbf{Losses:}
The normal maps $I_{gt}^n$, part segmentation maps $I_{gt}^s$, and depth maps $I_{gt}^z$ extracted from input images serve as pseudo-ground truth to optimize the reconstruction module. Instead of using normal maps, we use diffused maps $I_{gt}^d$ obtained by projecting light in the input camera view direction, for supervision. Thus, the rendering loss is defined as:
\begin{equation}
\mathcal{L}_{\text{diffuse}} = \mathcal{H}(( I_{gt}^s \odot I_\text{pred}^d), I_\text{gt}^d )   + \mathcal{S}(( I_\text{gt}^s \odot I_\text{pred}^d), I_\text{gt}^d) ),
\label{eq:l_diff}
\end{equation}

where $\odot$ is the element-wise multiplication, $\mathcal{H}$ is the Huber Loss, $\mathcal{S}$ is the SSIM loss and $I_\text{pred}^d$ is the diffuse images calculated from the input of the predicted mesh normals. 


The regularization loss, $ \mathcal{L}_{\text{reg}}$ ensures continuous surface consistency after deformation. The per-triangle Jacobians $J_j \in \mathbb{R}^{3 \times 3}$ of the final intrinsic field $J_t^F$ is optimized to be close to the identity matrix $I \in \mathbb{R}^{3 \times 3}$, defined as:
\begin{equation}
\mathcal{L}_{\text{reg}} = \sum_{j=1}^{M} \| J_j - I\|_2^2 .
\end{equation}

Finally, we use a depth supervision loss, $\mathcal{L}_{\text{depth}}$, calculated using the depth-ranking scheme proposed in \cite{sparsenerf}. A modified segmentation loss $\mathcal{L}_{\text{mask}}$ is used for supervision from segmentation masks (more detail in Suppl.). The total geometric reconstruction loss $\mathcal{L}_{\text{geo}}$ is defined as:
\begin{align}
    \mathcal{L}_{\text{geo}} = & \ \lambda_1 \mathcal{L}_{\text{render}} + \lambda_2 \mathcal{L}_{\text{mask}}  + \lambda_3 \mathcal{L}_{\text{reg}} + \lambda_4 \mathcal{L}_{\text{depth}}.
\end{align}

\input{images/fig_texture_architecture}

\subsubsection{\textbf{Gradient Based Adaptive Remeshing }}
\label{sec:remeshing}
 We select a set of edges $E_s$, based on their gradients from rendering loss $\mathcal{L}_{\text{diffuse}}$ and then apply remeshing operations, as illustrated in~\autoref{fig:remeshing}. 

\noindent \textbf{Edge Selection:} 
Out of all edges \(\mathbf{E}\) in the base mesh \(M^B\), we select a subset \(E_s \subset \mathbf{E}\) for remeshing. The image-space gradient at each pixel \( p\) is defined as \(\mathcal{G}(p) = \nabla_{I_{\text{pred}}^d(p)} \mathcal{L}_{\text{diffuse}}\) ~\autoref{eq:l_diff} where \( I_{\text{pred}}^d(p) \) is the predicted image. These pixel gradients are aggregated over rasterized faces \( \Pi(f_j) \) for each face \( f_j \in \mathbf{F} \), resulting in per rasterized face  gradient values \( \mathcal{G}(\Pi_{\text{raster}}(f_j)) \). These values are then aggregated over all iterations and projected onto the base mesh \(M^B\), yielding a per-face gradient value:  \(\mathcal{G}(f_j) = \frac{\Sigma \mathcal{G}(\Pi_{\text{raster}}(f_j))}{|\Pi_{\text{raster}}(f_j)|}\). Next, we select the top quantile of faces \(\mathcal{F}_{\omega} = \left\{ f_j \ \middle| \ \|\mathcal{G}(f_{j})\| \geq \text{quantile}_{\omega} \left( \|\mathcal{G}(f_{j})\| \right) \right\}\) for a percentile $\omega$ of triangle face. Subsequently, we prune all faces \( \mathcal{F}_{\delta} = \left\{ f_j \ \middle| \ L(e_l) \geq \delta_{\text{length}}, \ \forall e_l \in \mathcal{E}(f_j) \right\} \) whose edge lengths fall below a certain threshold $\delta_{\text{length}}$. The selection threshold and pruning criteria evolve over epochs over a linearly decaying function, ensuring a balance between preserving details and preventing excessive refinement. Finally, we select all edges $E_s$ part of all the final selected faces $\mathcal{F}_{\delta}$.


\noindent \textbf{Remeshing:} Subsequently, we perform edge splitting and edge flipping operations on \(E_s\), adopting the remeshing strategy proposed in~\cite{adaptiveisoremesh}.
During the remeshing process, it is crucial to handle face flips and degenerate triangles. Finally, we clean up the mesh to remove degenerate faces and merge close vertices. This yields the modified topology base mesh $M_r^B$. Next, we need to recomputation of all mesh attributes. After remeshing, the mesh attributes are recomputed via $k$-NN interpolation. The static Jacobian field $J^S$, Adam optimizer moments $m_1, m_2$, and skinning weights $W$ are interpolated to ensure smooth training. Please refer to Suppl. for more information. 


\input{images/fig_qualitative_geometry}

\subsection{Appearance Reconstruction Module} \label{subsec:texture_map}
Our goal is to learn a dynamic texture map corresponding to the reposed mesh at each frame. The detailed overview of texture recovery is provided in ~\autoref{fig:texture_archi}. 
We obtain the base UV coordinates $M_{UV}$ from  $M_r^B$ using~\cite{smartUVMap}. These UV coordinates map the color information from a texture map to the mesh faces. 
Similar to geometry reconstruction, we learn two texture components. A frame-invariant static texture map $T^S \in \mathbb{R}^{q \times q \times 3}$, and a per-frame dynamic texture map $T^D_t \in \mathbb{R}^{q \times q \times 3}$, where $q$ is the texture image dimension. The static texture $T^S$  is optimized directly, while the dynamic texture $T_t^D$ is predicted by a neural network. At every time-step $t$, the MLP $f_T$ is conditioned on hash encoded UV coordinates $f_\varphi(M_{UV})$, and pose parameters $\gamma(\theta_t)$ to predict $T^D_t$. The final Texture map is obtained as $T^F_t=T^S+T^D_t$. For better generalization, we employ a smooth annealing training strategy, inspired by \cite{deformgaussian}, wherein we introduce linearly decaying Gaussian noise to the pose parameters, $\gamma(\theta_t)$. This approach effectively mitigates overfitting and improves generalizability in novel views synthesis. At each iteration, the posed garment mesh $M_t^P$ from the geometry module is rendered with color from texture $T_t^{F}$, to produce colored images $I_{pred}^c$. The static texture $T^S$ and the neural network parameters of $f_T$ are optimized via differentiable rendering with the following two losses:
$\mathcal{L}_{\text{col}} = ||(I_{gt}^s \odot I_{pred}^c), I_{gt}^c||_1$ and 
$\mathcal{L}_{\text{ssim}} = \text{SSIM}(I_{gt}^s \odot I_{pred}^c), I_{gt}^c$. \\
The final loss is defined as:
\begin{equation}
    L_{tex}=\alpha_1 L_{col}+\alpha_2 L_{ssim}.
\end{equation}  

\input{images/fig_qualitative_peoplesnap}

%% file: images/fig_remeshing.tex
\begin{figure*}[t]
  \centering
   \includegraphics[width=\linewidth]{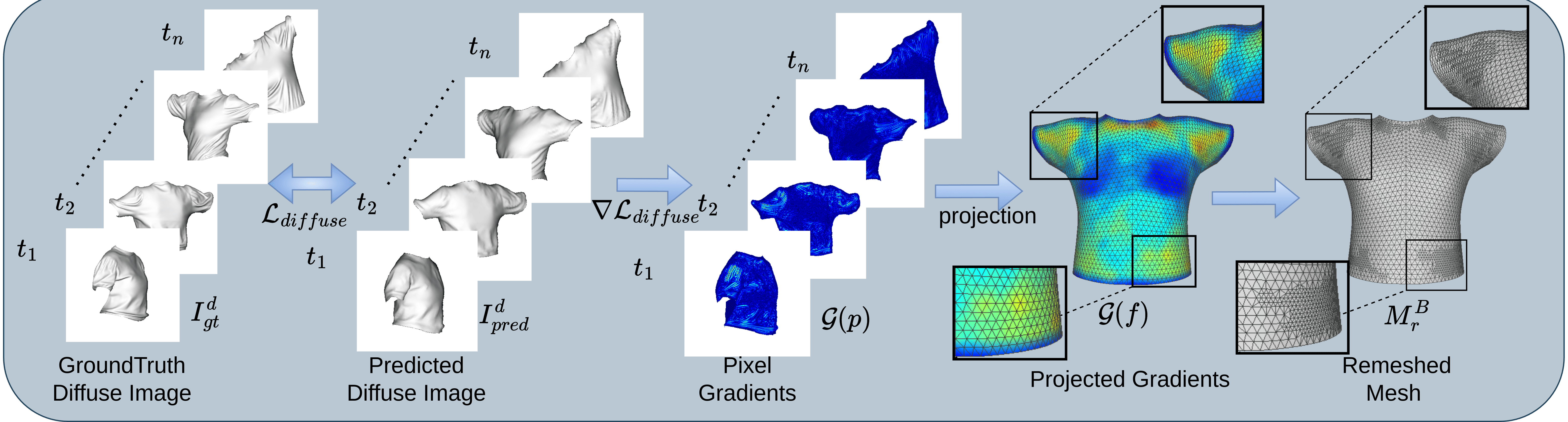}
   \caption{\textbf{Overview of our gradient-based adaptive remeshing method: }Performing edge selection, followed by remeshing operations for generating remeshed meshs with high frequency details.  }

\label{fig:remeshing}
\end{figure*}

%% file: images/fig_texture_architecture.tex
\begin{figure}
  \centering
    \includegraphics[width=\linewidth]{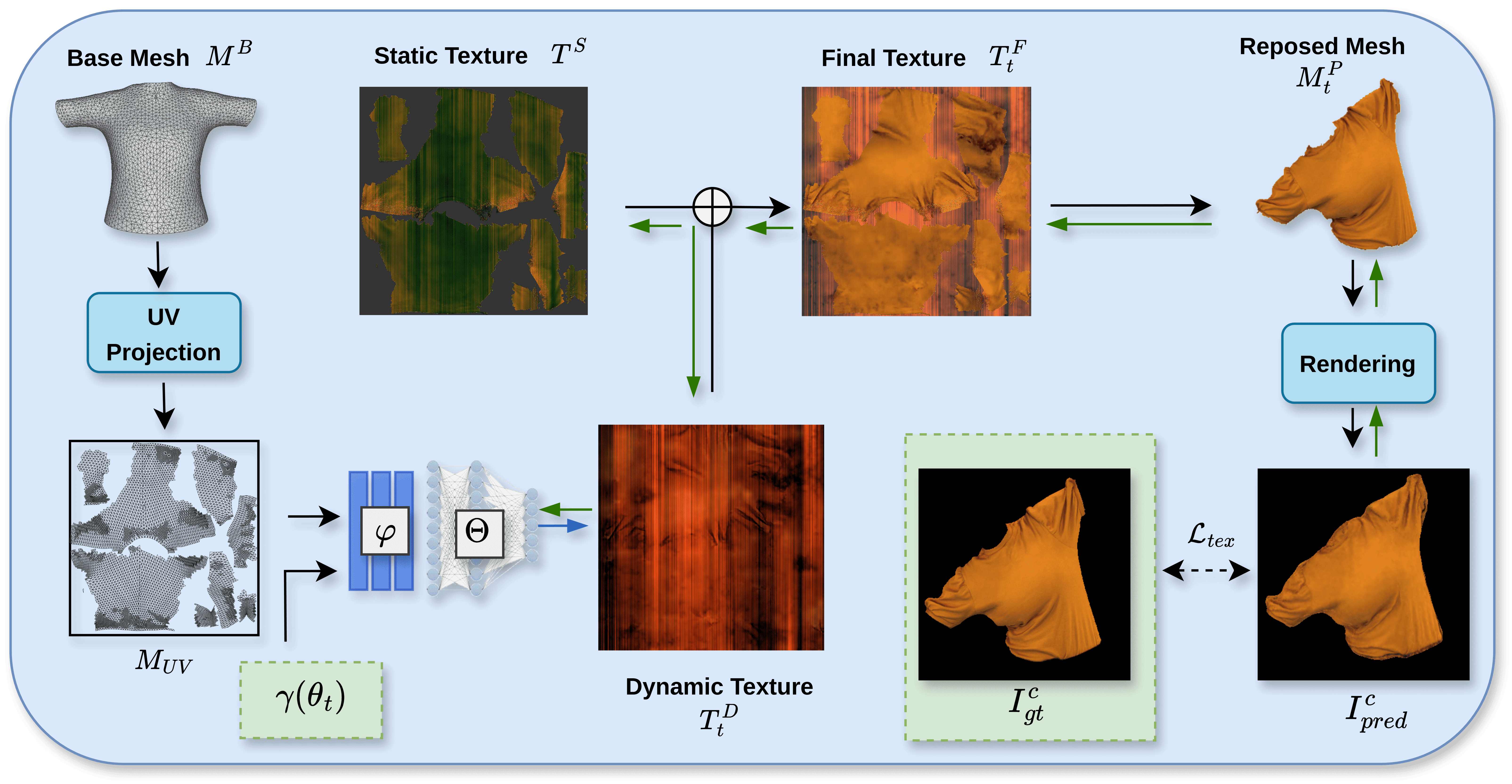}
    \caption{Overview of our appearance reconstruction module.}
    \label{fig:texture_archi}
\end{figure}

%% file: images/fig_qualitative_geometry.tex
\begin{figure*}
  \centering
    \includegraphics[width=\linewidth]{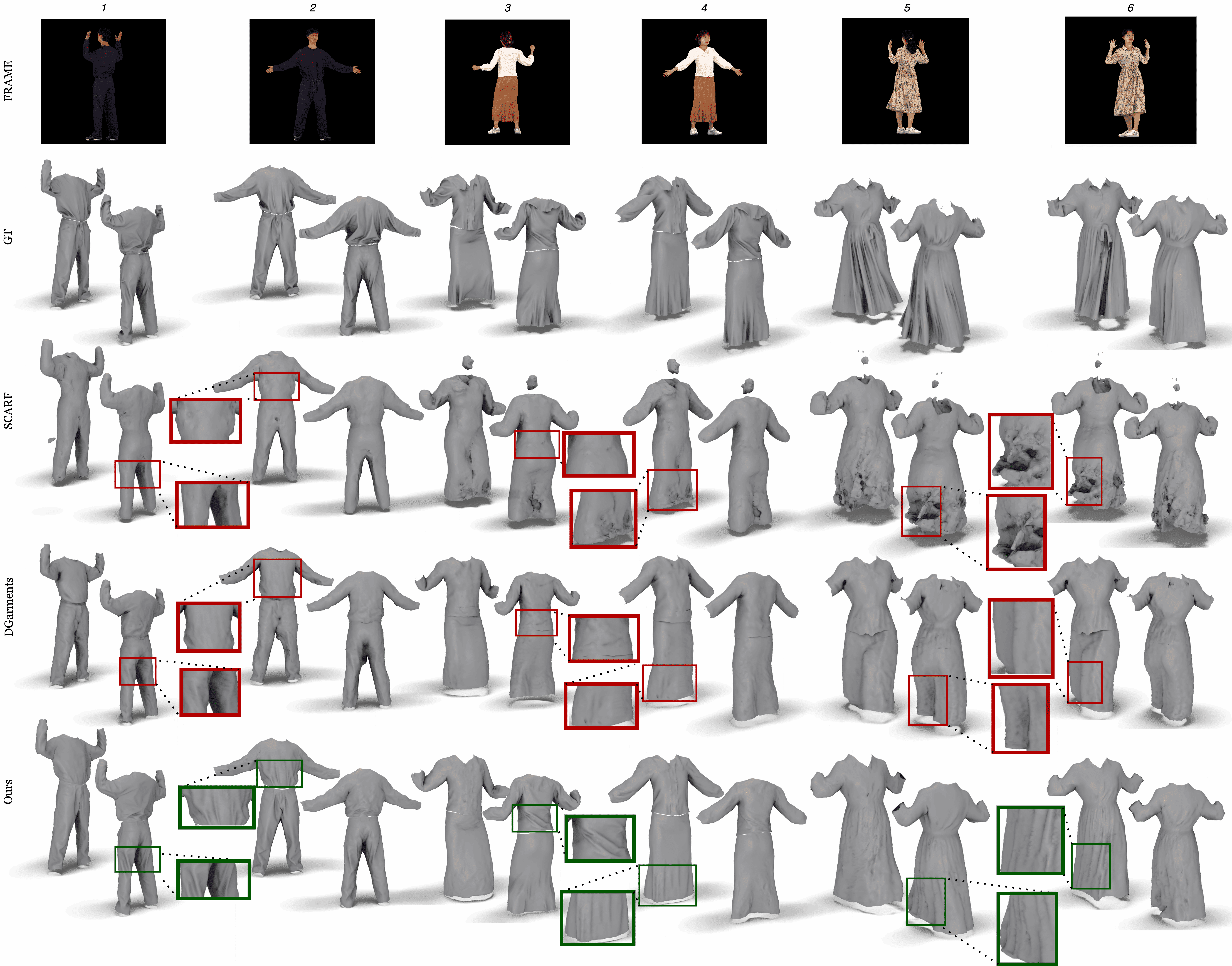}
    \caption{Qualitative comparison where  our method faithfully reconstructs high-frequency details like tiny wrinkles and folds, closer to GroundTruth in comparison to SCARF \cite{scarf} and DGarments \cite{dgarments} on 4D-Dress dataset \cite{4ddress}.}
    \label{fig:surface_compare_4ddress}
\end{figure*}

%% file: images/fig_qualitative_peoplesnap.tex
\begin{figure*}
  \centering
    \includegraphics[width=\linewidth]{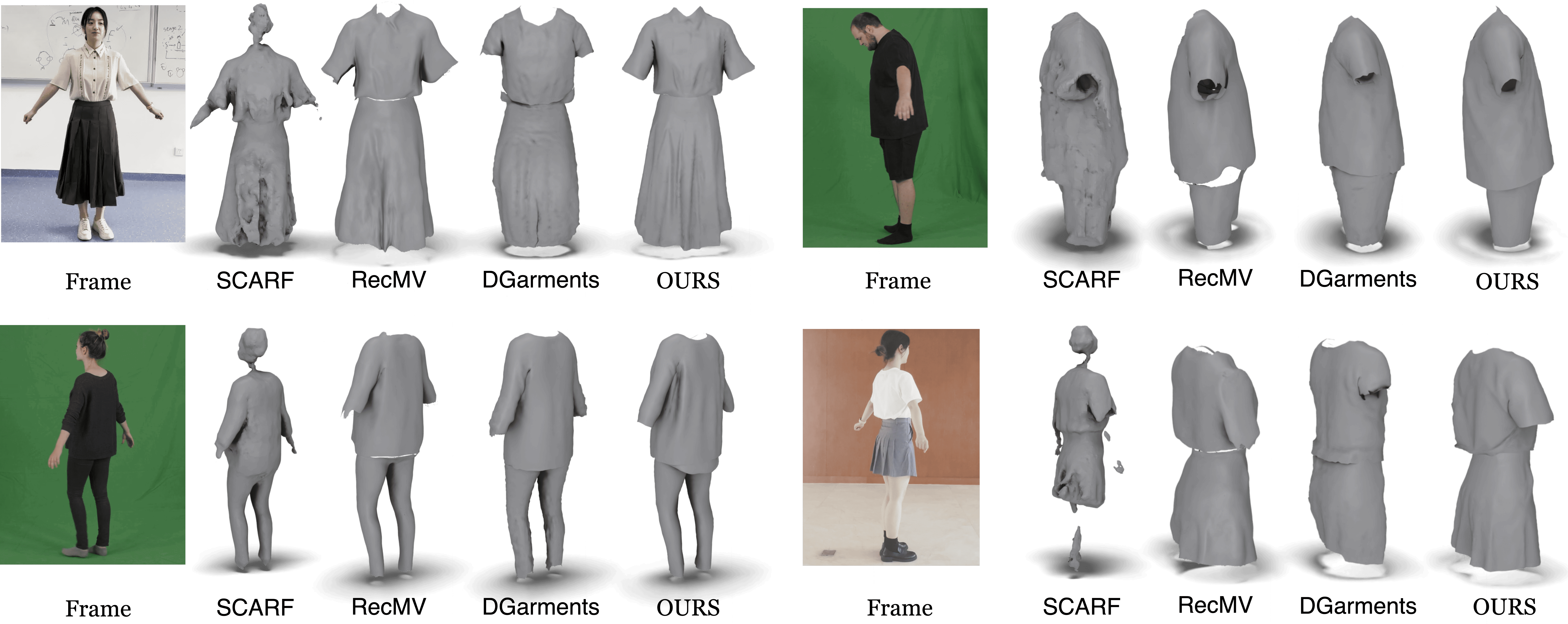}
    \caption{ Qualitative comparison of geometric reconstruction obtained by our method with SCARF \cite{scarf} and REC-MV \cite{recmv} on People Snapshot \cite{peoplesnapshot} dataset. Our method faithfully reconstructs high-frequency details like tiny wrinkles and folds.}
    \label{fig:surface_compare_people}
\end{figure*}

%% file: sec/4_Experiments.tex
\section{Experiments \& Results}
\label{sec:formatting}

\subsection{Implementation Details}
Our proposed method is implemented in PyTorch with NVDiffrast \cite{nvdiffrast} as the core differentiable rasterizer. The primary training for our method was conducted on a single NVIDIA RTX 4090 GPU, for both geometry and appearance reconstruction. Each sequence of 100 frames takes approximately 2.5 hours to train including texture recovery. Both modules incorporate a fixed-epoch warm-up phase, during which only the static deformation field $J^S$ and static texture map $T^S$ are optimized. After the warm-up phase, the dynamic deformation field $J_t^D$ and dynamic texture map $T_t^D$ are introduced for joint optimization. Adaptive remeshing is performed at fixed intervals throughout the optimization process.

\input{tables/table_quant_geometry}
\input{tables/table_quant_texture}

\subsection{Experimental Setup}
We evaluate and compare our method against recent State-Of-The-Art (SOTA) approaches on two tasks: 3D surface reconstruction and novel view synthesis. Our evaluation spans five sequences from a modified 4D-Dress \cite{4ddress} dataset, along with two additional datasets \cite{peoplesnapshot, recmv}, selecting two sequences from each to demonstrate robustness.  We provide quantitative comparisons for both tasks on the 4D-Dress dataset ~\cite{4ddress}. Additionally, we provide qualitative comparisons for 4D-Dress dataset for both tasks across all datasets. To assess the effectiveness of our model, we perform comparisons with the following SOTA methods - REC-MV \cite{recmv}, SCARF \cite{scarf}, and DGarment \cite{dgarments}. Finally, we provide extensive ablation studies to analyze our design choices. Please refer Suppl. for Dataset specifications and implementation details.
\input{images/fig_texture}
\input{tables/table_ablation}

\noindent \textbf{Data Preprocessing:} We utilize existing pre-trained vision models to obtain reliable priors. The SMPL pose and shape parameters and the camera estimations are obtained from 4DHumans \cite{4dhumans}.  Per-frame normal map, depth map, and part-segmentation are recovered using a pre-trained human foundation model Sapiens \cite{saipens}. Finally, the base garment mesh is obtained using BCNet \cite{bcnet}. 

\subsection{Results}
\noindent \textbf{Geometry Reconstruction :} Quantitative evaluation, presented in \autoref{tab:quantitative_novel} (rows [1-3]), demonstrates that our method significantly outperforms the SOTA methods \cite{scarf,dgarments} both in terms of Normal Consistency (NC) as well as Chamfer Distance (CD), averaged across all frames of a sequence. We achieve a significantly improved alignment of the reconstructed garment mesh with the ground truth mesh while achieving consistent geometrical characteristics across different frames, leading to substantially lower average CD values \& higher average NC values across the sequence as well as average overall sequences across the dataset. 

A similar trend is evident in the qualitative evaluation presented in \autoref{fig:surface_compare_4ddress}. The qualitative differences are more significant for col $[3-6]$ which contains loose clothing such as gown, where we outperform the existing methods while effectively mitigating major artifacts as shown in col $5$. The qualitative results for additional datasets \cite{peoplesnapshot, recmv} are shown in \autoref{fig:surface_compare_people} where we demonstrate our method's ability to preserve high-frequency details superior to other SOTA methods. 
Overall, due to the implicit nature of representation, both SCARF \cite{scarf} and REC-MV \cite{recmv} fail to capture high-fidelity details in the garments' geometry. However, DGarments \cite{dgarments} addresses this limitation by predicting a per-vertex displacement on the explicit mesh. Nevertheless, their method is unable to model large deformations and hence struggles to handle loose garments effectively. 


\textbf{Texture Reconstruction :} We present quantitative evaluations for novel view synthesis in \autoref{tab:quantitative_novel}, demonstrating that our method consistently outperforms the existing state-of-the-art across all visual evaluation metrics, including PSNR, SSIM~\cite{ssim}, and LPIPS~\cite{lpips}. 
This highlights the high fidelity and perceptual quality of our approach. Additionally, the qualitative comparisons in ~\autoref{fig:novel_compare_main} further reinforce the effectiveness of our method where in terms of the visual quality of the appearance, our method yields sharp textural details in comparison to SCARF \cite{scarf}.




%% file: tables/table_quant_geometry.tex
\setlength\dashlinedash{0.2pt}
\setlength\dashlinegap{3pt}
\setlength\arrayrulewidth{1.5pt}

\begin{table*}
  \centering
  \caption{Quantitative evaluation on geometry reconstruction on 4D-Dress dataset \cite{4ddress} using Chamfer Distance (CD) and Normal Consistency (NC) and comparison with different methods.}
  \begin{tabular}{lcccccccccccc}
    \toprule
    & \multicolumn{6}{c}{\textbf{Chamfer Distance} \( \mathcal{L}_2 \times 10^3 \downarrow\)} & \multicolumn{6}{c}{\textbf{Normal Consistency} \(\uparrow\)} \\
    \cmidrule(lr){2-7} \cmidrule(lr){8-13}
    Method & 123 & 148 & 169 & 185 & 187 & Avg & 123 & 148 & 169 & 185 & 187 & Avg \\
    \midrule
    SCARF       & 8.622  & -      & 6.507  & 2.423  & 3.261  & 5.203  & 0.915  & -      & 0.872  & 0.837  & 0.753  & 0.844 \\
    DGarment    & 0.076  & 0.863  & 0.154  & 0.431  & 1.722  & 0.649  & 0.904  & 0.755  & 0.872  & 0.856  & 0.777  & 0.833 \\

    \textbf{Ours}   & \textbf{0.050}  & \textbf{0.660}  & \textbf{0.127}  & 0.393  & \textbf{0.923}  & \textbf{0.431}  & \textbf{0.934}  & \textbf{0.766}  & \textbf{0.891}  & \textbf{0.879}  & \textbf{0.794}  & \textbf{0.853} \\
    \hdashline
    
    w/o remeshing & 0.053  & 0.672  & 0.129  & \textbf{0.372}  & 0.981  & 0.441  & 0.932  & 0.762  & 0.887  & 0.878  & 0.790  & 0.850 \\
 w normals & 0.195  & 0.931  & 0.278  & 0.535 & 1.205  & 0.554 & 0.908  & 0.755  & 0.866  & 0.853  & 0.778  & 0.832 \\
    \bottomrule
  \end{tabular}
  
  \label{tab:quantitative_surface}
\end{table*}

%% file: tables/table_quant_texture.tex
\begin{table*}
  \centering
  \caption{Quantitative evaluation on novel view synthesis with PSNR (PR), SSIM (SM), and LPIPS (LS) on different sequences.}
  \begin{tabular}{lcccccccccccc}
    \toprule
    Sequence & \multicolumn{3}{c}{123} & \multicolumn{3}{c}{169} & \multicolumn{3}{c}{185} & \multicolumn{3}{c}{187} \\
    \cmidrule(lr){2-4} \cmidrule(lr){5-7} \cmidrule(lr){8-10} \cmidrule(lr){11-13}
    Method & \textbf{PR} $\uparrow$ & \textbf{SM} $\uparrow$ & \textbf{LS} $\downarrow$ & \textbf{PR} $\uparrow$ & \textbf{SM} $\uparrow$ & \textbf{LS} $\downarrow$ & \textbf{PR} $\uparrow$ & \textbf{SM} $\uparrow$ & \textbf{LS} $\downarrow$ & \textbf{PR} $\uparrow$ & \textbf{SM} $\uparrow$ & \textbf{LS} $\downarrow$ \\
    \midrule
    SCARF & 43.02 & 0.992 & 0.018 & 45.01 & 0.992 & 0.026 & 33.82 & 0.986 & 0.025 & 25.32 & 0.918 & 0.0828 \\
    \textbf{Ours} & \textbf{46.78} & \textbf{0.998} & \textbf{0.008} & \textbf{47.91} & \textbf{0.996} & \textbf{0.014} & \textbf{35.21} & \textbf{0.990} & \textbf{0.017} & \textbf{25.85} & \textbf{0.948} & \textbf{0.0395} \\
    \bottomrule
  \end{tabular}
  
  \label{tab:quantitative_novel}
\end{table*}

%% file: images/fig_texture.tex
\begin{figure}
  \centering
    \includegraphics[width=\linewidth]{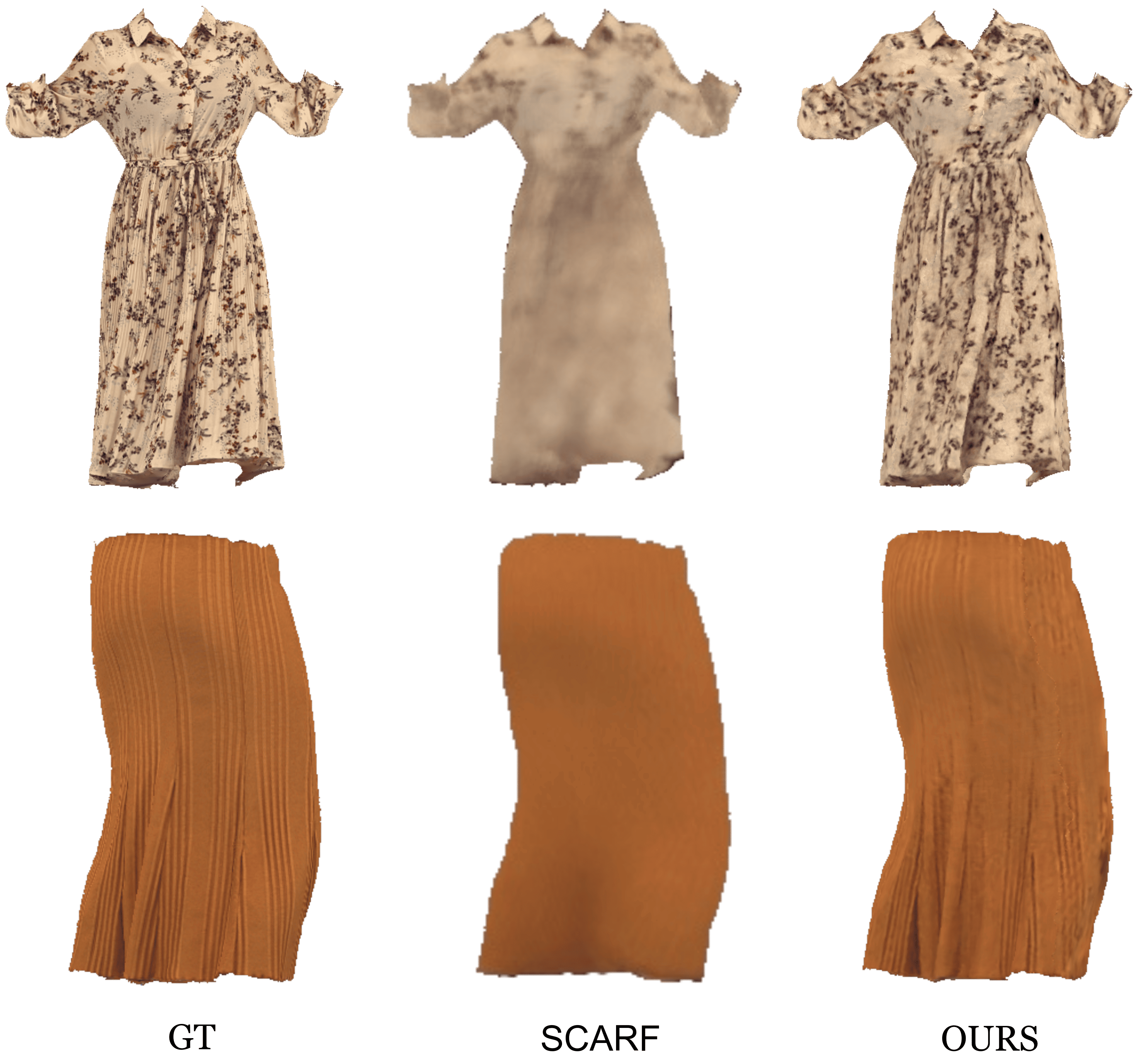}
    \caption{Qualitative comparison of novel view synthesis.}
    \label{fig:novel_compare_main}
\end{figure}

%% file: tables/table_ablation.tex

%% file: sec/5_Ablations.tex
\subsection{Ablation Studies}
\label{ablationstudies}


\input{images/fig_ablation_remeshing}
\textbf{Effect of Adaptive Remshing :} The effectiveness of our adaptive remeshing strategy is demonstrated in \autoref{tab:quantitative_surface} (rows [3,4]). Although the CD \& NC metrics show marginal quantitative degradation in case of without remeshing, we qualitatively demonstrate in~\autoref{fig:ablation_remeshing} that there is a significant drop in the fidelity of reconstructions (shown in row b), which is particularly leading to loss of complex folds and curved surfaces in comparison to reconstruction obtained with our full method (with remshing shown in row a). The remeshing process also effectively mitigates major artifacts by reducing the occurrence of larger triangles, as visible in the armpit region \autoref{fig:ablation_remeshing} (see red circle). Furthermore, our remeshing strategy adaptively increases resolution in regions with higher geometric variation, enabling more precise capture of details such as folds, pockets, and other fine cloth structures ( see \autoref{fig:ablation_remeshing} square box), resulting in more accurate reconstructions.

\input{images/fig_ablation_normal_diffuse}
\textbf{Normals vs Diffuse Image :} 
We observe that normals predicted (from~\cite{saipens}) in directions perpendicular to the viewing angles exhibit ambiguity. To address this limitation, we instead use diffuse images, which are basically the normals' components aligned with the viewing direction. Unlike standard normal maps, diffuse maps provide softer constraints, enabling improved generalization across frames. 
We provide empirical evaluation supporting the effectiveness of diffuse image supervision through ablation studies summarized in \autoref{tab:quantitative_surface} (rows [3,5]). Our results consistently demonstrate that incorporating diffuse image supervision leads to improved performance compared to normal image supervision, further validating this design choice. A similar trend is observed in the qualitative comparisons illustrated in \autoref{fig:ablation_normal}, where the use of diffuse images results in improved geometric detail compared to normal image supervision.  






%% file: images/fig_ablation_remeshing.tex
\begin{figure}
  \centering
    \includegraphics[width=\linewidth]{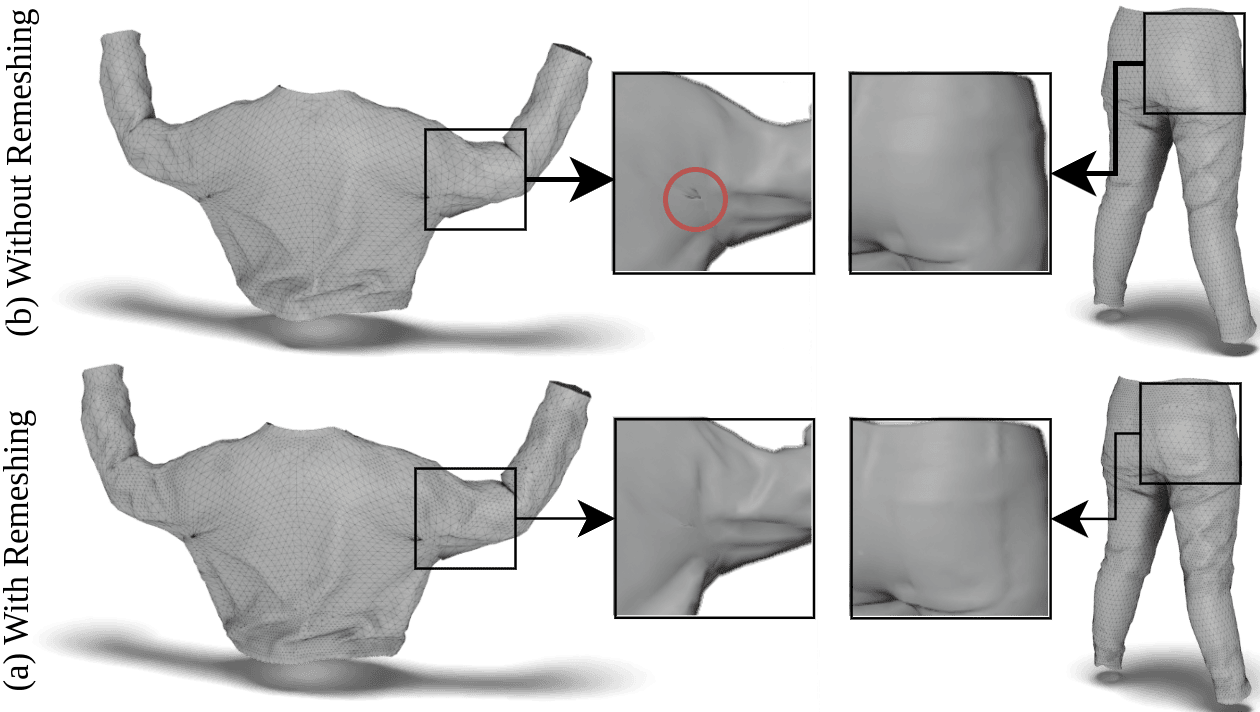}
    
    \caption{Ablative results on  Gradient Based Adaptive Remeshing.}
    \label{fig:ablation_remeshing}
\end{figure}

%% file: images/fig_ablation_normal_diffuse.tex
\begin{figure}
  \centering
    \includegraphics[width=\linewidth]{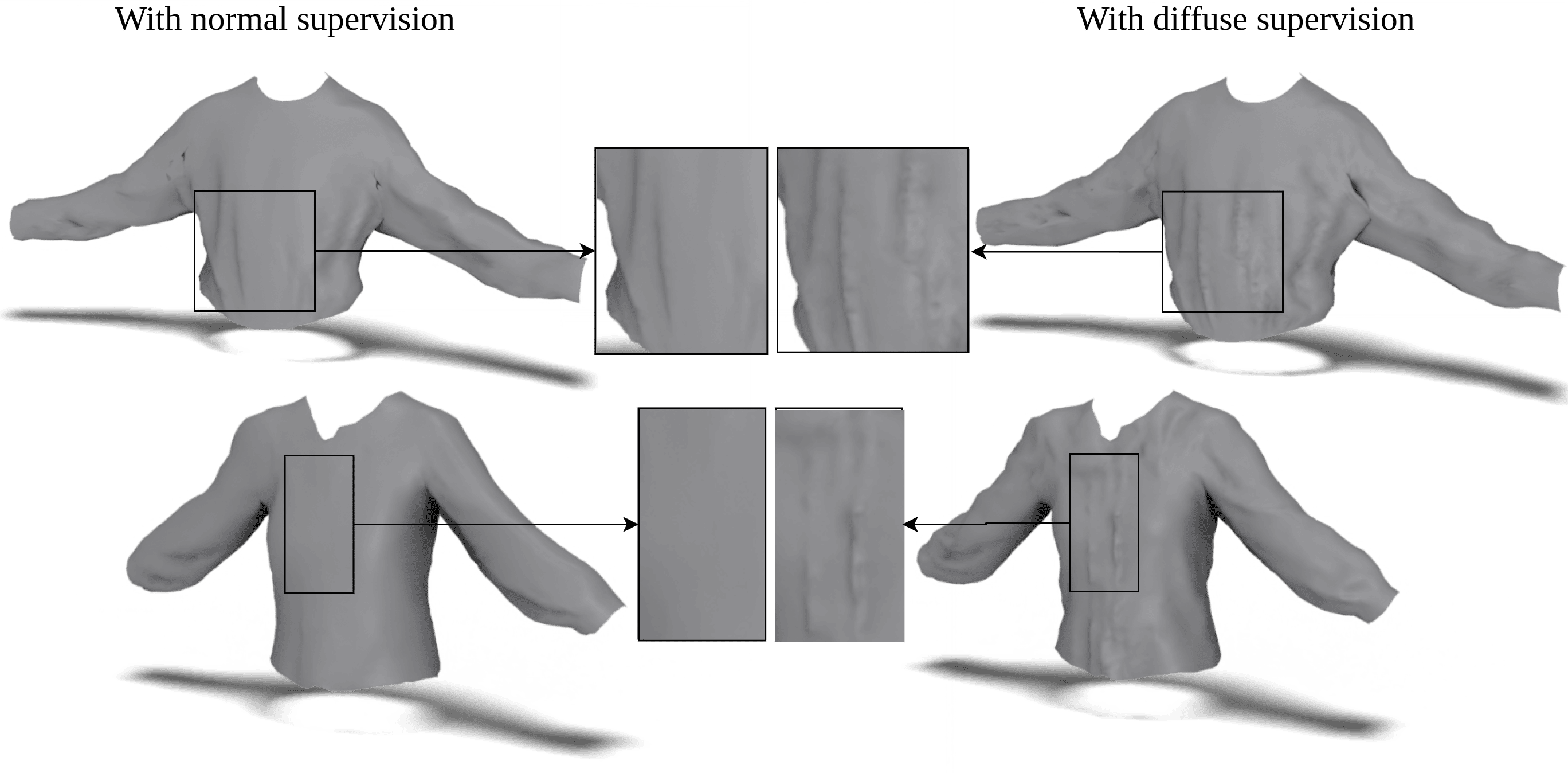}
    
    \caption{Ablative results comparing use of diffuse image supervision vs normal supervision.}
    \label{fig:ablation_normal}
\end{figure}

%% file: sec/6_Conclusion.tex
\section{Conclusion and Future Works}
\label{sec:formatting}

We propose a novel gradient-based deformation method to reconstruct dynamic textured garments from monocular video. We model both appearance and geometry, providing high-quality garment reconstruction. Our adaptive remeshing strategy facilitates modeling high-frequency details and extremely loose garments. We demonstrate the superiority of our method through improved qualitative and quantitative evaluations against SOTA. However, there is room for substantial improvement. One limitation of using a mesh representation instead of implicit functions is its susceptibility to self-intersection. A more robust method to actively prevent self-intersections could significantly enhance results. A more realistic solution would incorporate physics directly into the garment’s deformation representation, beyond simply adding it as a loss term.

\noindent \textbf{Acknowledgement:} We would like to thank Astitva Srivastava and Chandradeep Pokhariya for their helpful discussions and feedback, which greatly improved this work.

%% file: main.bbl
\begin{thebibliography}{51}
\providecommand{\natexlab}[1]{#1}
\providecommand{\url}[1]{\texttt{#1}}
\expandafter\ifx\csname urlstyle\endcsname\relax
  \providecommand{\doi}[1]{doi: #1}\else
  \providecommand{\doi}{doi: \begingroup \urlstyle{rm}\Url}\fi

\bibitem[Aigerman et~al.(2022)Aigerman, Gupta, Kim, Chaudhuri, Saito, and Groueix]{NJF}
Noam Aigerman, Kunal Gupta, Vladimir~G. Kim, Siddhartha Chaudhuri, Jun Saito, and Thibault Groueix.
\newblock Neural jacobian fields: learning intrinsic mappings of arbitrary meshes.
\newblock \emph{ACM Trans. Graph.}, 41\penalty0 (4), 2022.

\bibitem[Alldieck et~al.(2018)Alldieck, Magnor, Xu, Theobalt, and Pons-Moll]{peoplesnapshot}
Thiemo Alldieck, Marcus Magnor, Weipeng Xu, Christian Theobalt, and Gerard Pons-Moll.
\newblock Video based reconstruction of 3d people models.
\newblock In \emph{{IEEE}/{CVF} Conference on Computer Vision and Pattern Recognition ({CVPR})}, pages 8387--8397, 2018.
\newblock {CVPR} Spotlight Paper.

\bibitem[Alldieck et~al.(2022)Alldieck, Zanfir, and Sminchisescu]{PHORHUM}
Thiemo Alldieck, Mihai Zanfir, and Cristian Sminchisescu.
\newblock Phorhum, 2022.

\bibitem[Casado-Elvira et~al.(2022)Casado-Elvira, Trinidad, and Casas]{pergamo}
Andr{\'e}s Casado-Elvira, Marc~Comino Trinidad, and Dan Casas.
\newblock Pergamo: Personalized 3d garments from monocular video.
\newblock In \emph{Computer Graphics Forum}, pages 293--304. Wiley Online Library, 2022.

\bibitem[Chen et~al.(2023)Chen, Yang, Fu, Meng, Chen, Yang, and Gao]{implicitPCA}
Lan Chen, Jie Yang, Hongbo Fu, Xiaoxu Meng, Weikai Chen, Bo Yang, and Lin Gao.
\newblock Implicitpca: Implicitly-proxied parametric encoding for collision-aware garment reconstruction.
\newblock \emph{Graph. Models}, 129\penalty0 (C), 2023.

\bibitem[Corona et~al.(2021)Corona, Pumarola, Alenyà, Pons-Moll, and Moreno-Noguer]{smplicit}
Enric Corona, Albert Pumarola, Guillem Alenyà, Gerard Pons-Moll, and Francesc Moreno-Noguer.
\newblock Smplicit: Topology-aware generative model for clothed people, 2021.

\bibitem[Corona et~al.(2024)Corona, Alenyà, Pons-Moll, and Moreno-Noguer]{layernet}
Enric Corona, Guillem Alenyà, Gerard Pons-Moll, and Francesc Moreno-Noguer.
\newblock Layernet: High-resolution semantic 3d reconstruction of clothed people.
\newblock \emph{IEEE Transactions on Pattern Analysis and Machine Intelligence}, 46\penalty0 (2):\penalty0 1257--1272, 2024.

\bibitem[Dunyach et~al.(2013)Dunyach, Vanderhaeghe, Barthe, and Botsch]{adaptiveisoremesh}
Marion Dunyach, David Vanderhaeghe, Lo{\"\i}c Barthe, and Mario Botsch.
\newblock Adaptive remeshing for real-time mesh deformation.
\newblock In \emph{Eurographics 2013}. The Eurographics Association, 2013.

\bibitem[Feng et~al.(2022)Feng, Yang, Pollefeys, Black, and Bolkart]{scarf}
Yao Feng, Jinlong Yang, Marc Pollefeys, Michael~J. Black, and Timo Bolkart.
\newblock Capturing and animation of body and clothing from monocular video.
\newblock In \emph{SIGGRAPH Asia 2022 Conference Papers}, 2022.

\bibitem[Gao et~al.(2023)Gao, Aigerman, Groueix, Kim, and Hanocka]{textdeformer}
William Gao, Noam Aigerman, Thibault Groueix, Vova Kim, and Rana Hanocka.
\newblock Textdeformer: Geometry manipulation using text guidance.
\newblock In \emph{ACM SIGGRAPH 2023 Conference Proceedings}, pages 1--11, 2023.

\bibitem[Goel et~al.(2023)Goel, Pavlakos, Rajasegaran, Kanazawa, and Malik]{4dhumans}
Shubham Goel, Georgios Pavlakos, Jathushan Rajasegaran, Angjoo Kanazawa, and Jitendra Malik.
\newblock Humans in 4{D}: Reconstructing and tracking humans with transformers.
\newblock In \emph{ICCV}, 2023.

\bibitem[Guo et~al.(2023)Guo, Jiang, Chen, Song, and Hilliges]{vid2avatar}
Chen Guo, Tianjian Jiang, Xu Chen, Jie Song, and Otmar Hilliges.
\newblock Vid2avatar: 3d avatar reconstruction from videos in the wild via self-supervised scene decomposition, 2023.

\bibitem[Guo et~al.(2024)Guo, Jiang, Kaufmann, Zheng, Valentin, Song, and Hilliges]{reloo}
Chen Guo, Tianjian Jiang, Manuel Kaufmann, Chengwei Zheng, Julien Valentin, Jie Song, and Otmar Hilliges.
\newblock Reloo: Reconstructing humans dressed in loose garments from monocular video in the wild, 2024.

\bibitem[Habermann et~al.(2020)Habermann, Xu, Zollhofer, Pons-Moll, and Theobalt]{DeepCap}
Marc Habermann, Weipeng Xu, Michael Zollhofer, Gerard Pons-Moll, and Christian Theobalt.
\newblock Deepcap: Monocular human performance capture using weak supervision.
\newblock In \emph{Proceedings of the IEEE/CVF Conference on Computer Vision and Pattern Recognition}, pages 5052--5063, 2020.

\bibitem[He et~al.(2021)He, Xu, Saito, Soatto, and Tung]{arch++}
Tong He, Yuanlu Xu, Shunsuke Saito, Stefano Soatto, and Tony Tung.
\newblock Arch++: Animation-ready clothed human reconstruction revisited.
\newblock pages 11026--11036, 2021.

\bibitem[Ho et~al.(2024)Ho, Song, and Hilliges]{sith}
Hsuan-I Ho, Jie Song, and Otmar Hilliges.
\newblock Sith: Single-view textured human reconstruction with image-conditioned diffusion, 2024.

\bibitem[Huang et~al.(2020)Huang, Xu, Lassner, Li, and Tung]{ARCH}
Zeng Huang, Yuanlu Xu, Christoph Lassner, Hao Li, and Tony Tung.
\newblock Arch: Animatable reconstruction of clothed humans.
\newblock \emph{2020 IEEE/CVF Conference on Computer Vision and Pattern Recognition (CVPR)}, pages 3090--3099, 2020.

\bibitem[Jiang et~al.(2020)Jiang, Zhang, Hong, Luo, Liu, and Bao]{bcnet}
Boyi Jiang, Juyong Zhang, Yang Hong, Jinhao Luo, Ligang Liu, and Hujun Bao.
\newblock Bcnet: Learning body and cloth shape from a single image.
\newblock In \emph{Computer Vision – ECCV 2020: 16th European Conference, Glasgow, UK, August 23–28, 2020, Proceedings, Part XX}, page 18–35, Berlin, Heidelberg, 2020. Springer-Verlag.

\bibitem[Jiang et~al.(2022)Jiang, Hong, Bao, and Zhang]{selfrecon}
Boyi Jiang, Yang Hong, Hujun Bao, and Juyong Zhang.
\newblock Selfrecon: Self reconstruction your digital avatar from monocular video.
\newblock In \emph{Proceedings of the IEEE/CVF Conference on Computer Vision and Pattern Recognition}, pages 5605--5615, 2022.

\bibitem[Kanazawa et~al.(2018)Kanazawa, Tulsiani, Efros, and Malik]{icmr}
Angjoo Kanazawa, Shubham Tulsiani, Alexei~A. Efros, and Jitendra Malik.
\newblock Learning category-specific mesh reconstruction from image collections.
\newblock In \emph{ECCV}, 2018.

\bibitem[Kerbl et~al.(2023)Kerbl, Kopanas, Leimk{\"u}hler, and Drettakis]{gaussianSplatting}
Bernhard Kerbl, Georgios Kopanas, Thomas Leimk{\"u}hler, and George Drettakis.
\newblock 3d gaussian splatting for real-time radiance field rendering.
\newblock \emph{ACM Transactions on Graphics}, 42\penalty0 (4), 2023.

\bibitem[Khirodkar et~al.(2024)Khirodkar, Bagautdinov, Martinez, Zhaoen, James, Selednik, Anderson, and Saito]{saipens}
Rawal Khirodkar, Timur Bagautdinov, Julieta Martinez, Su Zhaoen, Austin James, Peter Selednik, Stuart Anderson, and Shunsuke Saito.
\newblock Sapiens: Foundation for human vision models.
\newblock \emph{arXiv preprint arXiv:2408.12569}, 2024.

\bibitem[Laine et~al.(2020)Laine, Hellsten, Karras, Seol, Lehtinen, and Aila]{nvdiffrast}
Samuli Laine, Janne Hellsten, Tero Karras, Yeongho Seol, Jaakko Lehtinen, and Timo Aila.
\newblock Modular primitives for high-performance differentiable rendering.
\newblock \emph{ACM Transactions on Graphics}, 39\penalty0 (6), 2020.

\bibitem[L\'{e}vy et~al.(2002)L\'{e}vy, Petitjean, Ray, and Maillot]{smartUVMap}
Bruno L\'{e}vy, Sylvain Petitjean, Nicolas Ray, and J\'{e}rome Maillot.
\newblock Least squares conformal maps for automatic texture atlas generation.
\newblock \emph{ACM Trans. Graph.}, 21\penalty0 (3):\penalty0 362–371, 2002.

\bibitem[Li et~al.(2022)Li, Guillard, Remelli, and Fua]{dig}
Ren Li, Benoît Guillard, Edoardo Remelli, and Pascal Fua.
\newblock Dig: Draping implicit garment over the human body, 2022.

\bibitem[Li et~al.(2024{\natexlab{a}})Li, Dumery, Guillard, and Fua]{li2024_garmentrecovery}
Ren Li, Corentin Dumery, Benoît Guillard, and Pascal Fua.
\newblock Garment recovery with shape and deformation priors, 2024{\natexlab{a}}.

\bibitem[Li et~al.(2023)Li, Zhang, Lai, Yang, and Li]{dgarments}
Xiongzheng Li, Jinsong Zhang, Yu-Kun Lai, Jingyu Yang, and Kun Li.
\newblock High-quality animatable dynamic garment reconstruction from monocular videos.
\newblock \emph{IEEE Transactions on Circuits and Systems for Video Technology}, 2023.

\bibitem[Li et~al.(2024{\natexlab{b}})Li, Chen, Larionov, Sarafianos, Matusik, and Stuyck]{diffavatar}
Yifei Li, Hsiao-yu Chen, Egor Larionov, Nikolaos Sarafianos, Wojciech Matusik, and Tuur Stuyck.
\newblock {DiffAvatar}: Simulation-ready garment optimization with differentiable simulation.
\newblock In \emph{Proceedings of the IEEE/CVF Conference on Computer Vision and Pattern Recognition (CVPR)}, 2024{\natexlab{b}}.

\bibitem[Liu et~al.(2023)Liu, Xu, Lin, Liang, and Yan]{sewformer}
Lijuan Liu, Xiangyu Xu, Zhijie Lin, Jiabin Liang, and Shuicheng Yan.
\newblock Towards garment sewing pattern reconstruction from a single image.
\newblock \emph{ACM Transactions on Graphics (SIGGRAPH Asia)}, 2023.

\bibitem[Luigi et~al.(2023)Luigi, Li, Guillard, Salzmann, and Fua]{drapenet}
Luca~De Luigi, Ren Li, Benoît Guillard, Mathieu Salzmann, and Pascal Fua.
\newblock Drapenet: Garment generation and self-supervised draping, 2023.

\bibitem[Mildenhall et~al.(2021)Mildenhall, Srinivasan, Tancik, Barron, Ramamoorthi, and Ng]{nerf}
Ben Mildenhall, Pratul~P Srinivasan, Matthew Tancik, Jonathan~T Barron, Ravi Ramamoorthi, and Ren Ng.
\newblock Nerf: Representing scenes as neural radiance fields for view synthesis.
\newblock \emph{Communications of the ACM}, 65\penalty0 (1):\penalty0 99--106, 2021.

\bibitem[Moon et~al.(2022)Moon, Nam, Shiratori, and Lee]{clothwild}
Gyeongsik Moon, Hyeongjin Nam, Takaaki Shiratori, and Kyoung~Mu Lee.
\newblock 3d clothed human reconstruction in the wild.
\newblock In \emph{Computer Vision – ECCV 2022: 17th European Conference, Tel Aviv, Israel, October 23–27, 2022, Proceedings, Part II}, page 184–200, Berlin, Heidelberg, 2022. Springer-Verlag.

\bibitem[Peng et~al.(2021)Peng, Zhang, Xu, Wang, Shuai, Bao, and Zhou]{neural_body}
Sida Peng, Yuanqing Zhang, Yinghao Xu, Qianqian Wang, Qing Shuai, Hujun Bao, and Xiaowei Zhou.
\newblock Neural body: Implicit neural representations with structured latent codes for novel view synthesis of dynamic humans, 2021.

\bibitem[Qiu et~al.(2023)Qiu, Chen, Zhou, Xu, Wang, and Han]{recmv}
Lingteng Qiu, Guanying Chen, Jiapeng Zhou, Mutian Xu, Junle Wang, and Xiaoguang Han.
\newblock Rec-mv: Reconstructing 3d dynamic cloth from monocular videos.
\newblock In \emph{Proceedings of the IEEE/CVF Conference on Computer Vision and Pattern Recognition}, pages 4637--4646, 2023.

\bibitem[Rong et~al.(2024)Rong, Grigorev, Wang, Black, Thomaszewski, Tsalicoglou, and Hilliges]{GaussianGarments}
Boxiang Rong, Artur Grigorev, Wenbo Wang, Michael~J Black, Bernhard Thomaszewski, Christina Tsalicoglou, and Otmar Hilliges.
\newblock Gaussian garments: Reconstructing simulation-ready clothing with photorealistic appearance from multi-view video.
\newblock \emph{arXiv preprint arXiv:2409.08189}, 2024.

\bibitem[Saito et~al.(2020)Saito, Simon, Saragih, and Joo]{pifuhd}
Shunsuke Saito, Tomas Simon, Jason Saragih, and Hanbyul Joo.
\newblock Pifuhd: Multi-level pixel-aligned implicit function for high-resolution 3d human digitization.
\newblock In \emph{Proceedings of the IEEE Conference on Computer Vision and Pattern Recognition}, 2020.

\bibitem[Saito et~al.(2021)Saito, Yang, Ma, and Black]{scanimate}
Shunsuke Saito, Jinlong Yang, Qianli Ma, and Michael~J. Black.
\newblock {SCANimate}: Weakly supervised learning of skinned clothed avatar networks.
\newblock In \emph{Proceedings IEEE/CVF Conf.~on Computer Vision and Pattern Recognition (CVPR)}, 2021.

\bibitem[Srivastava et~al.(2025)Srivastava, Manu, Raj, Jampani, and Sharma]{wordrope}
Astitva Srivastava, Pranav Manu, Amit Raj, Varun Jampani, and Avinash Sharma.
\newblock Wordrobe: Text-guided generation of textured 3d garments.
\newblock In \emph{European Conference on Computer Vision}, pages 458--475. Springer, 2025.

\bibitem[Wang et~al.(2023)Wang, Chen, Loy, and Liu]{sparsenerf}
Guangcong Wang, Zhaoxi Chen, Chen~Change Loy, and Ziwei Liu.
\newblock Sparsenerf: Distilling depth ranking for few-shot novel view synthesis.
\newblock In \emph{Proceedings of the IEEE/CVF International Conference on Computer Vision}, pages 9065--9076, 2023.

\bibitem[Wang et~al.(2024)Wang, Ho, Guo, Rong, Grigorev, Song, Zarate, and Hilliges]{4ddress}
Wenbo Wang, Hsuan-I Ho, Chen Guo, Boxiang Rong, Artur Grigorev, Jie Song, Juan~Jose Zarate, and Otmar Hilliges.
\newblock 4d-dress: A 4d dataset of real-world human clothing with semantic annotations.
\newblock In \emph{Proceedings of the IEEE Conference on Computer Vision and Pattern Recognition (CVPR)}, 2024.

\bibitem[Wang et~al.(2004)Wang, Bovik, Sheikh, and Simoncelli]{ssim}
Zhou Wang, A.C. Bovik, H.R. Sheikh, and E.P. Simoncelli.
\newblock Image quality assessment: from error visibility to structural similarity.
\newblock \emph{IEEE Transactions on Image Processing}, 13\penalty0 (4):\penalty0 600--612, 2004.

\bibitem[Xiang et~al.(2020)Xiang, Prada, Wu, and Hodgins]{MonoClothCap}
Donglai Xiang, Fabian Prada, Chenglei Wu, and Jessica Hodgins.
\newblock Monoclothcap: Towards temporally coherent clothing capture from monocular rgb video, 2020.

\bibitem[Xiu et~al.(2022)Xiu, Yang, Tzionas, and Black]{icon}
Yuliang Xiu, Jinlong Yang, Dimitrios Tzionas, and Michael~J. Black.
\newblock {ICON}: {I}mplicit {C}lothed humans {O}btained from {N}ormals.
\newblock In \emph{Proceedings of the IEEE/CVF Conference on Computer Vision and Pattern Recognition (CVPR)}, pages 13296--13306, 2022.

\bibitem[Xiu et~al.(2023)Xiu, Yang, Cao, Tzionas, and Black]{ECON}
Yuliang Xiu, Jinlong Yang, Xu Cao, Dimitrios Tzionas, and Michael~J. Black.
\newblock Econ: Explicit clothed humans optimized via normal integration, 2023.

\bibitem[Yang et~al.(2024)Yang, Gao, Zhou, Jiao, Zhang, and Jin]{deformgaussian}
Ziyi Yang, Xinyu Gao, Wen Zhou, Shaohui Jiao, Yuqing Zhang, and Xiaogang Jin.
\newblock Deformable 3d gaussians for high-fidelity monocular dynamic scene reconstruction.
\newblock In \emph{Proceedings of the IEEE/CVF conference on computer vision and pattern recognition}, pages 20331--20341, 2024.

\bibitem[Zhang et~al.(2018)Zhang, Isola, Efros, Shechtman, and Wang]{lpips}
Richard Zhang, Phillip Isola, Alexei~A Efros, Eli Shechtman, and Oliver Wang.
\newblock The unreasonable effectiveness of deep features as a perceptual metric.
\newblock In \emph{Proceedings of the IEEE conference on computer vision and pattern recognition}, pages 586--595, 2018.

\bibitem[Zhang et~al.(2024)Zhang, Yang, and Yang]{sifu}
Zechuan Zhang, Zongxin Yang, and Yi Yang.
\newblock Sifu: Side-view conditioned implicit function for real-world usable clothed human reconstruction, 2024.

\bibitem[Zheng et~al.(2024)Zheng, Zhao, Yang, Yifan, Xiang, Dubost, Lagun, Beeler, Tombari, Guibas, et~al.]{physavatars}
Yang Zheng, Qingqing Zhao, Guandao Yang, Wang Yifan, Donglai Xiang, Florian Dubost, Dmitry Lagun, Thabo Beeler, Federico Tombari, Leonidas Guibas, et~al.
\newblock Physavatar: Learning the physics of dressed 3d avatars from visual observations.
\newblock \emph{arXiv preprint arXiv:2404.04421}, 2024.

\bibitem[Zheng et~al.(2020)Zheng, Yu, Liu, and Dai]{pamir}
Zerong Zheng, Tao Yu, Yebin Liu, and Qionghai Dai.
\newblock Pamir: Parametric model-conditioned implicit representation for image-based human reconstruction, 2020.

\bibitem[Zhu et~al.(2022)Zhu, Qiu, Qiu, and Han]{reef}
Heming Zhu, Lingteng Qiu, Yuda Qiu, and Xiaoguang Han.
\newblock Registering explicit to implicit: Towards high-fidelity garment mesh reconstruction from single images.
\newblock In \emph{Proceedings of the IEEE/CVF Conference on Computer Vision and Pattern Recognition (CVPR)}, pages 3845--3854, 2022.

\bibitem[Zielonka et~al.(2023)Zielonka, Bagautdinov, Saito, Zollh{\"o}fer, Thies, and Romero]{D3GA}
Wojciech Zielonka, Timur Bagautdinov, Shunsuke Saito, Michael Zollh{\"o}fer, Justus Thies, and Javier Romero.
\newblock Drivable 3d gaussian avatars.
\newblock \emph{arXiv preprint arXiv:2311.08581}, 2023.

\end{thebibliography}
